\documentclass{article}

\usepackage{arxiv}

\usepackage[utf8]{inputenc} % allow utf-8 input
\usepackage[T1]{fontenc}    % use 8-bit T1 fonts
\usepackage{hyperref}       % hyperlinks
\usepackage{url}            % simple URL typesetting
\usepackage{booktabs}       % professional-quality tables
\usepackage{amsfonts}       % blackboard math symbols
\usepackage{nicefrac}       % compact symbols for 1/2, etc.
\usepackage{microtype}      % microtypography
\usepackage{lipsum}
\usepackage{graphicx}
\usepackage{multirow}
\usepackage{color, colortbl}
\usepackage{authblk}
\definecolor{lightgray}{rgb}{0.83, 0.83, 0.83}
\graphicspath{ {./images/} }

\title{Biomedical and Clinical Language Models for Spanish: On the Benefits of Domain-Specific Pretraining in a Mid-Resource Scenario}

\author[1]{Casimiro Pio Carrino}
\author[1]{Jordi Armengol-Estap\'e}
\author[1]{Asier Gutiérrez-Fandiño}
\author[1]{Joan Llop-Palao}
\author[1]{Marc Pàmies}
\author[1]{Aitor Gonzalez-Agirre}
\author[1]{Marta Villegas}
\affil[1]{Text Mining Unit\\
Barcelona Supercomputing Center
}
\affil[ ]{\textit {\{casimiro.carrino,jordi.armengol,asier.gutierrez,joan.lloppalao,marc.pamies,aitor.gonzalez,marta.villegas\}@bsc.es}}

\begin{document}
\maketitle
\begin{abstract}
This work presents biomedical and clinical language models for Spanish by experimenting with different pretraining choices, such as masking at word and subword level, varying the vocabulary size and testing with domain data, looking for better language representations. Interestingly, in the absence of enough clinical data to train a model from scratch, we applied mixed-domain pretraining and cross-domain transfer approaches to generate a performant bio-clinical model suitable for real-world clinical data. We evaluated our models on Named Entity Recognition (NER) tasks for biomedical documents and challenging hospital discharge reports. When compared against the competitive mBERT and BETO models, we outperform them in all NER tasks by a significant margin. Finally, we studied the impact of the model’s vocabulary on the NER performances by offering an interesting vocabulary-centric analysis. The results confirm that domain-specific pretraining is fundamental to achieving higher performances in downstream NER tasks, even within a mid-resource scenario. To the best of our knowledge, we provide the first biomedical and clinical transformer-based pretrained language models for Spanish, intending to boost native Spanish NLP applications in biomedicine. Our best models are freely available in the HuggingFace hub: \url{https://huggingface.co/BSC-TeMU}.

\end{abstract}

\section{Introduction}
The success of transformer-based models in the general domain \cite{devlin-etal-2019-bert} has encouraged the development of language models for domain-specific scenarios \cite{chalkidis-etal-2020-legal, tai-etal-2020-exbert, DBLP:journals/corr/abs-1908-10063, DBLP:journals/corr/abs-1906-02124}. Specifically, in the biomedical domain, there has been a proliferation of models \cite{peng-etal-2019-transfer, beltagy-etal-2019-scibert,alsentzer-etal-2019-publicly, pubmedbert} since the first BioBERT \cite{10.1093/bioinformatics/btz682} model was published. Unfortunately, there is still a significant lack of biomedical and clinical models in languages other than English, despite the increasing efforts of the NLP community \cite{Nvol2014ClinicalNL, schneider-etal-2020-biobertpt}. Consequently, general-domain pretrained language models supporting Spanish, such as mBERT \cite{devlin-etal-2019-bert} and BETO \cite{beto}, are often used as a proxy in the absence of a genuine domain-specialized model. To fill this gap, we trained several biomedical language models experimenting with a wide range of pretraining choices, namely, masking with word and subword units, varying the vocabulary size and experimenting with the data domain. Furthermore, we applied cross-domain transfer and mixed-domain pretraining using biomedical and clinical data to train bio-clinical models intended for clinical applications. To evaluate our models, we choose the fundamental Named Entity Recognition (NER) task in information retrieval defined for biomedical and clinical scenarios, the former based on biomedical documents and the latter on real hospital documents. As the main result, our models obtained a significant gain over both mBERT and BETO models in all tasks. Additionally, we studied the impact of the model’s vocabulary on several downstream tasks by performing a vocabulary-centric analysis. The evaluation results highlights the importance of domain-specific pretraining over continual pretraining from general domain data in a mid-resource scenario. Our main contributions are:

\begin{enumerate}
    \item We release the first Spanish biomedical and clinical transformer-based pretrained language models, trained with the largest biomedical corpus known to date.
    \item We assess the effectiveness of these models in different settings, two biomedical scenarios and a demanding clinical scenario with real hospital reports.
    \item We show that biomedical models exhibit a remarkable cross-domain transfer ability on the clinical domain.
    \item We perform an in-depth vocabulary and segmentation analysis, which offers insights on domain-specific pretraining and raises interesting open questions.
\end{enumerate}

\section{Related work}
In the last years, several language models for both the biomedical and clinical domain have been trained through unsupervised pretraining of transformer-based architectures \cite{kalyan2021ammu}. The first trained model was BioBERT \cite{10.1093/bioinformatics/btz682}, where the authors adapted the BERT model \cite{devlin-etal-2019-bert}, trained with general-domain data, to the biomedical domain by continual pretraining.  Similarly, other works followed the continual pretraining approach to train the BlueBERT \cite{peng-etal-2019-transfer} and ClinicalBERT \cite{alsentzer-etal-2019-publicly} models. When enough in-domain data is available, training from scratch has been used as an alternative method to continual pretraining, leading to the SciBERT \cite{beltagy-etal-2019-scibert} and PubMedBERT \cite{pubmedbert} models. However, the SciBERT model uses mixed-domain data from the biomedical and computer science domain, while PubMedBERT leverages only data belonging to the biomedical domain. Interestingly, in \cite{pubmedbert}, the authors call into question the benefits of mixed-domain pretraining, estimating its negative impact on a set of downstream tasks belonging to an extensive biomedical benchmark (named BLURB) that they provide.

Our pretraining approach employed training from scratch with mixed-domain data by combining biomedical and clinical resources. However, in contrast to \cite{beltagy-etal-2019-scibert, pubmedbert} i) we used a corpus 3-5 times smaller, ii) we performed pretraining from scratch with mixed-domain data from biomedical documents and clinical notes and iii) we assess the suitability of cross-domain transfer from the  biomedical to the clinical domain.

\section{Corpora}\label{sec:corpora}
This work considers two corpora with very different sizes and domains, namely, a clinical corpus and a biomedical one. The \textbf{clinical corpus} contains 91M tokens from more than 278K clinical documents (including discharge reports, clinical course notes and X-ray reports). For the \textbf{biomedical corpus} we gathered  data from a variety of sources, namely:

% a miscellany of medical content, essentially clinical cases\footnote{A clinical case report is a type of scientific publication where medical practitioners share patient cases.}; scientific literature from Scielo and PubMed; medical patents; a Wikipedia health crawler; the EMEA corpus; the Spanish content from Medline; the background corpus from BARR2 Shared Task \cite{barr2_corpus} and a massive Spanish health domain crawled data. The crawling was conducted during the year 2020 on more than 3,000 Spanish domains associated to the biomedical domain, related to medical societies, scientific societies, journals, research centers, pharmaceutical companies, health educational websites, patient associations, personal web pages from healthcare professionals, hospital websites, medical regulatory colleges, as well as healthcare institutions and organizations. 
\begin{itemize}
    \item Medical crawler\footnotemark[3]: Crawler of more than 3,000 URLs belonging to Spanish biomedical and health domains introduced in \cite{carrino2021spanish}.
    \item Clinical cases misc.: A miscellany of medical content, essentially clinical case. Note that a clinical case report is different from a scientific publication where medical practitioners share patient cases and it is different from a clinical note or document.
    \item Scielo\footnotemark[4]: Scientific publications written in Spanish crawled from the Spanish SciELO server in 2017.
    \item PubMed: Open-access articles from the PubMed repository crawled in 2017.
    \item BARR2\_background\footnotemark[5]: Biomedical Abbreviation Recognition and Resolution (BARR2) containing Spanish clinical case study sections from a variety of clinical disciplines.
    \item Wikipedia\_life\_sciences: Wikipedia articles crawled on 04/01/2021 with the Wikipedia API python library\footnote{\url{https://pypi.org/project/Wikipedia-API/}} starting from the "Ciencias\_de\_la\_vida" category up to a maximum of 5 subcategories. Multiple links to the same articles are then discarded to avoid repeating content.
    \item Patents: Google Patent in Medical Domain for Spain (Spanish). The accepted codes (Medical Domain) for Json files of patents are: "A61B", "A61C","A61F", "A61H", "A61K", "A61L","A61M", "A61B", "A61P".
    \item EMEA\footnotemark[6]: Spanish-side documents extracted from parallel corpora made out of PDF documents from the European Medicines Agency.
    \item mespen\_Medline\footnotemark[7]: Spanish-side articles extracted from a collection of Spanish-English parallel corpus consisting of biomedical scientific literature.  The collection of parallel resources are aggregated from the MedlinePlus source.
\end{itemize}
Notice that, although the biomedical documents undoubtedly share a significant percentage of medical terms with clinical notes, the syntax and vocabulary may change radically due to the specific contexts and the idiosyncrasies of the user-generated content in clinical texts.

We cleaned each biomedical corpus to get the final corpus, and left the clinical corpus uncleaned. For each biomedical resource, we applied a cleaning pipeline with customized operations designed to read data in different formats, split it into sentences, detect the language, remove noisy and ill-formed sentences, deduplicate and eventually output the data with their original document boundaries. Finally, to remove repetitive content, we concatenated the entire corpus and deduplicate again, obtaining a total of 968M words. Table \ref{biomedical-corpus} shows detailed statistics of each dataset.

\begin{table*}
    \centering
    \begin{tabular}{l|r|r|r}
    \toprule
        \textbf{Corpus name}  & \textbf{No. tokens} & \textbf{Documents} & \textbf{Sentences} \\ \hline
        Medical crawler & 745,705,946 & 1,580,577 & 34,347,553 \\
        Clinical cases misc. & 102,855,267 & 41,592 & 5,565,854 \\
        Scielo & 60,007,289 & 16,981 & 2,668,231 \\
        BARR2\_background  & 24,516,442 & 127,635 & 1,029,600 \\
        Wikipedia\_life\_sciences & 13,890,501 & 15,791 & 832,027 \\
        Patents & 13,463,387 & 99,611 & 253,924 \\
        EMEA  & 5,377,448 & 284,119 & 284,575 \\
        mespen\_Medline  & 4,166,077 & 5,293 & 322,619 \\
        PubMed & 1,858,966 & 892 & 103,674 \\
         \bottomrule
    \end{tabular}
    \caption{List of individual sources constituting the biomedical corpus. The number of tokens refers to \emph{white-spaced} tokens calculated on cleaned untokenized text.}
    \label{biomedical-corpus}
\end{table*}

\section{Models pretraining}
The models trained in this work employed a RoBERTa \cite{DBLP:journals/corr/abs-1907-11692} base model with 12 self-attention layers. Following the original training, we prescind the auxiliary Next Sentence Prediction task used in BERT, and used masked language modelling as the pretraining objective. We used Subword Masking (SWM) as in \cite{DBLP:journals/corr/abs-1907-11692}, and the Whole Word Masking (WWM) \footnote{\url{https://github.com/google-research/bert} technique that masks all sub-words belonging to the same word \cite{Cui2019PreTrainingWW}.}.

We tokenized with the Byte-Level BPE algorithm introduced in \cite{radford2019language} and employed in the original RoBERTa \cite{DBLP:journals/corr/abs-1907-11692}, unlike previous biomedical language models \cite{beltagy-etal-2019-scibert, pubmedbert, 10.1093/bioinformatics/btz682} that use WordPiece \cite{devlin-etal-2019-bert} or SentencePiece \cite{kudo-2018-subword} segmentations.
We learned a cased vocabulary of 52k and 30k tokens to perform a comparative analysis. 

\footnotetext[3]{\url{https://zenodo.org/record/4561970}}
\footnotetext[4]{https://github.com/PlanTL-SANIDAD/SciELO-Spain-Crawler}
\footnotetext[5]{\url{https://temu.bsc.es/BARR2/downloads/background_set.raw_text.tar.bz2}}
\footnotetext[6]{\url{http://opus.nlpl.eu/download.php?f=EMEA/v3/moses/en-es.txt.zip}}
\footnotetext[7]{\url{https://zenodo.org/record/3562536}}
\stepcounter{footnote}
\stepcounter{footnote}
\stepcounter{footnote}

\stepcounter{footnote}

\stepcounter{footnote}

We run the training for 48 hours with 16 NVIDIA V100 GPUs of 16GB DDRAM, using Adam optimizer \cite{Adam} with peak learning rate of 0.0005 and an effective batch size of 2,048 sentences\footnote{Through gradient accumulation as implemented in Fairseq \cite{ott-etal-2019-fairseq}.}. We left the other hyperparameters in their default values as the original RoBERTa trainings. We then selected the model with the lowest perplexity in a holdout subset as the best model. Moreover, training was performed at the document level, preserving document boundaries. Document-level training may be crucial to enforce long-range dependencies and push the model towards the comprehension of entire documents, fostering the modelling of long-range dependencies.

We applied the pretraining method described above to train a variety of models, which can be divided into two groups, the Biomedical language models and the Bio-clinical language models.

\paragraph{Biomedical language models:}
We used the biomedical corpora described in section \ref{sec:corpora} of about 968M tokens to train biomedical language models. To study the impact of the masking mechanism and the vocabulary size, we experimented with the SWM and WWM techniques and with vocabularies of 52k and 30k tokens. We refer to the four variants as bio-52k-SWM, bio-52k-WWM, bio-30k-SWM and bio-30k-WWM. 

\paragraph{Bio-clinical language models:}
Due to the lack of a large-scale clinical corpus of comparable size to the biomedical one, we combined the biomedical and clinical corpora described in section \ref{sec:corpora} to train bio-clinical language models suitable for clinical settings. Furthermore, we also trained a bio-clinical model variant that leverages a vocabulary learned only from the clinical corpus with a vocabulary size of 52k. We refer to these two models as bio-cli-52k and bio-cli-52k-vocab-cli.

\section{Downstream NER tasks} \label{sec:ner_tasks}
We performed Named Entity Recognition (NER) tasks as a testbed for the our models since they are essential building blocks for many biomedical Text Mining and NLP applications. We employed a standard linear layer as a token classification head and the BIO tagging schema \cite{sang2000introduction} to fine-tune the pretrained models for the NER tasks. During fine-tuning, both the pretrained model and the classification layers parameters are learned with stochastic gradient descent.

Fine-tuning pretrained transformer-based language models for NER tasks by adding a linear layer on top of them is a usual practice in the literature (both in general-purpose models \cite{devlin-etal-2019-bert, DBLP:journals/corr/abs-1907-11692} and domain-specific ones \cite{10.1093/bioinformatics/btz682}). Remarkably, this method obtained impressive performances compared to more sophisticated classification layers such as Conditional Random Field layers on top of Bidirectional Long Short-Term Memory Recurrent Networks \cite{panchendrarajan-amaresan-2018-bidirectional}. Furthermore, its simplicity allows a head-to-head comparison of different pretraining strategies and baseline models, emphasising the ability of the pretrained representations.

We applied the fine-tuning method described above to three different NER datasets. The first two are data from two shared tasks and use annotations on curated medical data (clinical cases extracted from medical literature). The last one uses medical records from the ICTUSnet project\footnote{\url{https://ictusnet-sudoe.eu/es/}}.

\textbf{PharmaCoNER} \cite{gonzalez-agirre-etal-2019-pharmaconer} is a track on chemical and drug mention recognition from Spanish medical texts. The authors collected a manually classified collection of clinical case report sections derived from open access Spanish medical publications, named the Spanish Clinical Case Corpus (SPACCC). The corpus contained a total of 1,000 clinical cases and 396,988 words and was manually annotated, with a total of 7,624 entity mentions, corresponding to four different mention types\footnote{For a detailed description, see \url{https://temu.bsc.es/pharmaconer/}}. The track received several system submissions from the NLP community \cite{stoeckel-etal-2019-specialization, Akhtyamova2020TestingCW, 9087359, xiong-etal-2019-deep}

\textbf{CANTEMIST} \cite{miranda2020named} is a shared task specifically focusing on named entity recognition of tumor morphology, in Spanish. The CANTEMIST corpus\footnote{CANTEMIST corpus: \url{https://doi.org/10.5281/zenodo.3878178}} is a collection of 1,301 oncological case reports written in Spanish, with a total of 63,016 sentences and 1,093,501 tokens.
Several systems employing different strategies have been proposed to tackle the task \cite{vicomtech-cantemist,Xiong-cantemist, Vunikili-cantemist}

The \textbf{ICTUSnet} data set consists of 1,006 hospital discharge reports of patients admitted for stroke from 18 different Spanish hospitals. It contains more than 79,000 annotations for 51 different kinds of variables. The dataset is part of the ICTUSnet project, whose main objective was the development of an information extraction system to support domain experts when identifying relevant information in discharge reports.

\section{Evaluation and Main Results}\label{sec:eval_and_results}
%To assess the ability of our models, we performed the evaluation on the previous NER datasets. Due to the different data domain, we actually performed an evaluation for two domains, i) a biomedical domain, where task data are annotated over curated medical sources, belonging to the CANTEMIST and PharmaCoNER test sets and ii) a clinical domain, where annotations are extracted directly from hospital discharge reports, belonging to the ICTUSnet test set. Therefore, for each NER dataset and model, we fine-tune for the NER task for 10 epochs with a batch size of 32 and a maximum sentence length of 512 tokens. After training, we select the model with the maximum F1 score on the dev set as the best model. Finally, we computed the evaluation scores on the test set with the best model, as reported in table \ref{tab:evaluation-results}. \textcolor{yellow}{mencionar optimizador}
We evaluated our models on two biomedical benchmarks, PharmaCoNER and CANTEMIST, as well as the clinical ICTUSnet dataset, described in section \ref{sec:ner_tasks}. For each NER dataset and model, we fine-tune for 10 epochs with a batch size of 32 and a maximum sentence length of 512 tokens. After training, we select as the best model the one with the highest F1 score on the development set. Finally, we computed the evaluation scores by feeding the test set to the best model. Tables \ref{tab:biomedical_eval} and \ref{tab:bioclinical_eval} show the results for the biomedical models and bio-clinical models, respectively. We would like to remark that the ICTUSnet dataset is a challenging task since its consists of real hospital discharge reports. Moreover, for the biomedical models, the ICTUSnet evaluation represents a cross-domain transfer experiment from the biomedical to clinical domain.

Overall, our models achieved the best scores, beating both mBERT and BETO significantly. The biomedical models showed a remarkable cross-domain transfer ability compared to the bio-clinical models, achieving competitive performances on the ICTUSnet task. Nonetheless, the bio-clinical models obtained the best performances, indicating that the mixed-domain pretraining is practical approach to mitigate the lack of large-scale real-world clinical data. These results confirm the suitability of pretraining from scratch in a scenario with medium-size resources. We finally point out that, although our primary objective aside from achieving the best task performance, the results obtained are promising, evincing that a more sophisticated classification layer could further improve the tasks performances.

\begin{table*}
\centering
\begin{tabular}{l|cc|cc|c|c}
\hline
\toprule
 & \multicolumn{4}{c|}{\textbf{Bio}} &  \textbf{mBERT} & \textbf{BETO} \\ \hline
vocab size &  \multicolumn{2}{c|}{52k} & \multicolumn{2}{c|}{30k} & 120k & 31k \\
masking & SWM & WWM & SWM & WWM & SWM & WWM\\\hline
% & \textbf{bio-SWM-52k} & \textbf{bio-SWM-30k} & \textbf{bio-WWM-52k} & \textbf{bio-WWM-30k}
\textbf{PharmaCoNER} &  &  & & & &  \\
\hspace{1mm}- F1 & 89.48$\pm$ 0.60 & 89.62$\pm$0.57 & 89.47$\pm$0.33 & \textbf{89.85$\pm$0.47} & 87.46$\pm$0.23 & 88.18$\pm$0.41\\\hline
\hspace{1mm}- Precision & 87.85$\pm$ 1.15 & 88.18$\pm$0.99 & 88.33$\pm$0.42 & 88.41$\pm$0.41 & 86.50$\pm$0.95 & 87.12$\pm$0.52  \\
\hspace{1mm}- Recall  & 91.18$\pm$ 0.74 & 91.11$\pm$0.40 & 90.64$\pm$0.76 & 91.35$\pm$0.58 & 88.46$\pm$0.57 & 89.28$\pm$0.69 \\ 
\hline\hline
\textbf{CANTEMIST} & & & & & & \\
\hspace{1mm}- F1 & \textbf{83.87$\pm$ 0.41} & 83.00$\pm$0.17 & 82.85$\pm$0.36 & 83.23$\pm$0.34 & 82.61$\pm$0.67 & 82.42$\pm$0.06\\\hline
\hspace{1mm}- Precision & 81.70$\pm$0.52 & 80.81$\pm$0.27 & 80.93$\pm$0.48 & 81.27$\pm$0.72 & 81.12$\pm$0.65 & 80.91$\pm$0.41  \\
\hspace{1mm}- Recall  & 86.17$\pm$0.46 & 85.32$\pm$0.40 & 84.87$\pm$0.27 & 85.29$\pm$0.52 & 84.15$\pm$0.78 & 84.00$\pm$0.45 \\ 
\hline\hline
\rowcolor{lightgray}\textbf{ICTUSnet} &  &  & & & &  \\
\rowcolor{lightgray}\hspace{1mm}- F1 & 88.12$\pm$0.25 & 87.94$\pm$0.43 & 87.77$\pm$0.35 & \textbf{88.21$\pm$0.26} & 86.75$\pm$0.21  & 85.95$\pm$0.25\\\hline
\rowcolor{lightgray}\hspace{1mm}- Precision & 85.56$\pm$0.26 & 84.88$\pm$0.80 & 84.33$\pm$0.28 & 85.20$\pm$0.57 & 83.53$\pm$0.35 & 83.10$\pm$0.57  \\
\rowcolor{lightgray}\hspace{1mm}- Recall  & 90.83$\pm$0.47 & 91.22$\pm$0.49 & 91.50$\pm$0.66 & 91.45$\pm$0.44 & 90.23$\pm$0.46 & 89.02$\pm$0.61 \\ 
\bottomrule
\end{tabular}
\caption{Evaluation scores (F1, Precision and Recall) for the PharmaCoNER, CANTEMIST and ICTUSnet NER tasks. We compared our biomedical models with two general-domain baseline models, namely, multilingual BERT and BETO. Rows in light gray stress that the Biomedical model is performing cross-domain transfer of the ICTUSnet task since it belongs to the clinical domain. The scores are averaged across 5 random runs with standard deviation as error bar.}
\label{tab:biomedical_eval}
\end{table*}

\begin{table*}
\centering
\begin{tabular}{l|cc|cc}
\hline
\toprule
 &  \multicolumn{2}{|c|}{\textbf{Bio-cli}} &  \textbf{mBERT} & \textbf{BETO} \\ \hline
 vocab size-domain &  52k & 52k-vocab-cli & 120k-general & 31k-general \\ \hline
\textbf{PharmaCoNER} &  &  &   \\
\hspace{1mm}- F1 & \textbf{90.04$\pm$0.13} & 89.29$\pm$0.52 & 87.46$\pm$0.23 & 88.18$\pm$0.41 \\ \hline
\hspace{1mm}- Precision & 88.92$\pm$0.36 & 87.67$\pm$0.96 & 86.50$\pm$0.95 & 87.12$\pm$0.52   \\
\hspace{1mm}- Recall  & 91.18$\pm$0.29 & 90.98$\pm$0.47 & 88.46$\pm$0.57 & 89.28$\pm$0.69  \\ 
\hline\hline
\textbf{CANTEMIST} &  &  & &    \\
\hspace{1mm}- F1 & 83.34$\pm$0.39 & \textbf{84.41$\pm$0.33} & 82.61$\pm$0.67 & 82.42$\pm$0.06 \\\hline
\hspace{1mm}- Precision & 81.48$\pm$0.66 & 82.75$\pm$0.25 & 81.12$\pm$0.65 & 80.91$\pm$0.41   \\
\hspace{1mm}- Recall  & 85.30$\pm$0.45 & 86.15$\pm$0.53 & 84.15$\pm$0.78 & 84.00$\pm$0.45  \\ 
\hline\hline
\textbf{ICTUSnet} &  &  &    \\
\hspace{1mm}- F1 & 88.08$\pm$0.12 & \textbf{88.45$\pm$0.63} &  86.75$\pm$0.21  & 85.95$\pm$0.25 \\\hline
\hspace{1mm}- Precision & 84.92$\pm$0.29 & 84.92$\pm$0.86 & 83.53$\pm$0.35 & 83.10$\pm$0.57 \\
\hspace{1mm}- Recall  & 91.50$\pm$0.29 & 92.3$\pm$0.52 & 90.23$\pm$0.46 & 89.02$\pm$0.61  \\ 
 \bottomrule
\end{tabular}
\caption{Evaluation scores (F1, Precision and Recall) for the PharmaCoNER, CANTEMIST and ICTUSnet NER tasks. We compared our Bio-clinical models with two baselines obtained with non domain-specific models in Spanish, namely, multilingual BERT and BETO. The scores are averaged across 5 random runs with standard deviation as error bar.}
\label{tab:bioclinical_eval}
\end{table*}

\section{Discussion and Analysis}
This section attempts to shed light on the evaluation results by conducting a vocabulary analysis and segmentation experiments.
\subsection{Vocabulary overlap}
We look at the evaluation results through the lens of the model's vocabulary. Undoubtedly, the vocabulary plays a crucial role during the downstream transfer since it is responsible for encoding the task-specific data.  Intuitively, we expect that the more the overlap between vocabulary and task's tokens, the better. A high overlap leverages more pretrained representations that could be beneficial for fine-tuning. Specifically, we hypothesize that the performance on each downstream task may be related to the number of vocabulary tokens used to encode the task's data. Therefore, we first tokenize the three NER tasks with each model's vocabulary and then calculate the vocabulary overlap with each task, expressed as the number of tokens. The results shown in Table \ref{tab:vocab_overlap_tasks} seem to support our hypothesis by showing that the best evaluation scores for each task are obtained by the model with the maximum number of tokens overlap. Accordingly, the lowest overlap exhibited by the baselines models may explain their lowest evaluation scores. 
\begin{table*}[ht]
\centering
\begin{tabular}{l|cc|cc}
\toprule
\textbf{Model} & \textbf{PharmaCoNER} & \textbf{CANTEMIST}  & \textbf{ICTUSnet}   \\ \hline
Bio-cli-52k  & \textbf{20,620 (40\%)} & 22,001 (42\%) & 23,360 (45\%) \\
Bio-cli-vocab-cli-52k   & 20,335 (39\%) & \textbf{23,095 (44\%)} & \textbf{30,467 (59\%)} \\
Bio-52k & 19,978 (38\%) & 20,951 (40\%) & 21,449 (41\%)\\
Bio-30k  & 15,792 (53\%) & 16,302 (54\%) & 16,266 (54\%)\\
\hline \hline
BETO    & 12,829 (41\%) & 13,044 (42\%) & 13,388 (43\%)  \\
mBERT   & 11,084 (9\%) & 11,434 (9\%) & 13,187 (11\%) \\
\bottomrule
\end{tabular}
\caption{The vocabulary overlap is expressed as the intersection of each model's vocabulary and segmented tokens for each NER task. We also show the percentage of overlapped tokens to the total vocabulary size. Note that the SWM and WWM variants of the biomedical models are merged under the same names since they are equivalent for the vocabulary analysis.}
\label{tab:vocab_overlap_tasks}
\end{table*}

\subsection{Impact of segmentation}
Intuitively, it is reasonable to assume that a proper domain-specific segmentation should preserve the integrity of biomedical terms, minimizing the number of subword units required to encode them. As pointed out in \cite{pubmedbert}, models employing out-of-domain vocabulary "\textit{are forced to divert parametrization capacity and training bandwidth to model biomedical terms using fragmented subwords"}. Therefore, from one side, over-segmenting may have a negative influence on the downstream performance. On the other side, under-segmentation that employs term-specific units could dramatically increase the vocabulary size. Moreover, under-segmenting could prevent the model from detecting relatedness between terms based on the shared subwords, especially in morphologically rich terminology. Therefore, we conducted a meticulous analysis of domain-specific terms segmentation to study the trade-off between under and over-segmentation conditions and shed light on its relation with the NER downstream performances.

\subsubsection{Splitting terms}
We analyzed, both qualitatively and quantitatively, how the different models under study segment biomedical terms. We retrieved all the biomedical terms used as annotations in the CANTEMIST, PharmaCoNER and ICTUSnet tasks and segmented them applying each model's tokenizer. Table \ref{segmentation-examples} shows the quality of segmentation on a random set of NER annotations, making a comparison between mBERT, BETO and our best model bio-cli-52k-vocab-cli. Table \ref{segmentation-stats} shows the average number of subwords generated by each model computed over all the NER annotations. As expected, mBERT and BETO split terms into many more pieces than our models, on average. However, the variations across biomedical and clinical models also point out that the vocabulary size and mixed-domain pretraining are relevant factors determining the final segmentation. Finally, Table \ref{segmentations-percentages} illustrates, for each model, how many tokens are segmented with a given number of subwords (up to more than 4 subwords). Again, both mBERT and BETO tend to over-segment when compared to our models. As an example, our best model (bio-cli-vocab-cli-52k) has roughly 50\% of PharmaCoNER annotations segmented in either one or two tokens. In comparison, mBERT and BETO have less than 20\% of the annotations segmented in that same amount of subwords.

\subsubsection{Dissecting the F1 score}
\begin{figure}[ht]
    \centering
    \includegraphics[scale=0.525]{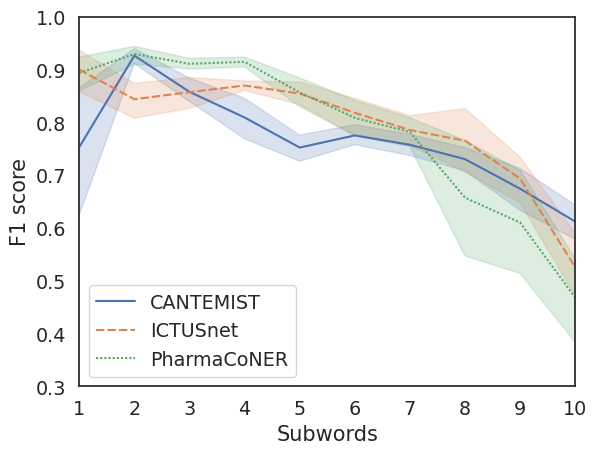}
    \caption{F1 score computed on the subset of annotations from the CANTEMIST, PharmaCoNer and ICTUSnet test sets that are shuttered in a given number of subword units.}
    \label{f1_factored}
\end{figure}

Finally, we seek a more precise relationship between the evaluation scores and the model's over-segmentation, expressed as the average number of subwords per term. Specifically, for each model and NER task, we group the annotations in the test split by the number of subwords they are shattered into and then recalculate the performance scores for each group of annotations. From the results presented in Figure \ref{f1_factored}, it can be observed that the F1 score decreases as the number of subwords increases. On average, the decreasing trend is exhibited across the three tasks, with a more considerable variation starting from 7 subwords. The analysis suggests that over-segmentation should be avoided in order to obtain higher F1 scores, confirming the intuition that a helpful segmentation preserves terms integrity. Note that, rather than criticising subword segmentation, we believe that further experiments unravelling the relationship between the segmentation and the downstream task performances could guide the design of an optimal biomedical segmentation.

% \addtolength{\tabcolsep}{-4pt}
% \begin{tabular}{lcc}
% \toprule
% \textbf{Models pairs} & \multicolumn{1}{l}{} & \multicolumn{1}{c}{\textbf{Median}} \\ \midrule
% bio-cli-52k & bio-52k & 0.8435 \\
% bio-30k & bio-52k & 0.5768 \\
% bio-cli-52k & bio 30k & 0.5566 \\
% bio-cli-52k & bio-cli-vocab-cli-52k & 0.3205 \\
% bio-52k & bio-cli-vocab-cli-52k & 0.2778 \\
% bio-30k & bio-cli-vocab-cli-52k & 0.2583 \\
%  \bottomrule
% \end{tabular}
% \addtolength{\tabcolsep}{4pt}
% \caption{Vocabulary overlapping between models. Note that the subwords and whole words masking variants of the biomedical models are merged under the same names since they are equivalent for the vocabulary analysis.}
% \label{tab:my-table}
% \end{table}

\begin{table}[ht]
\centering
\addtolength{\tabcolsep}{1pt}
\begin{tabular}{lcc}
\toprule
\textbf{Model} & \multicolumn{1}{l}{\textbf{Mean}} & \multicolumn{1}{c}{\textbf{Median}} \\ \midrule
bio-cli-vocab-cli-52k & 6.01 & 4 \\
bio-cli-52k           & 6.34 & 4 \\
bio-52k               & 6.63 & 5 \\
bio-30k               & 7.12 & 5 \\ \hline \hline
BETO                  & 8.60 & 6 \\
mBERT                 & 8.93 & 7 \\
 \bottomrule
\end{tabular}
\addtolength{\tabcolsep}{-1pt}
\caption{Comparison of mean and median number of tokens generated by each model on the PharmaCoNER, CANTEMIST and ICTUSnet annotations. Note that the SWM and WWM variants of the biomedical models are merged under the same names since they are equivalent for the vocabulary analysis.}
\label{segmentation-stats}
\end{table}

\begin{table*}
  \centering
  \begin{tabular}{rr|r|r|r|r|r|r}
  \toprule
    % \cline{3-8}
    \multicolumn{2}{c|}{}  & \textbf{mBERT} & \textbf{BETO} & \textbf{bio-30k} & \textbf{bio-52k} & \textbf{bio-cli-52k} & \textbf{bio-cli-vocab-cli-52k} \\\hline
  \multirow{3}{*}{{PharmaCoNER}} 
  & 1 tok  & 2.79\%  &  3.04\% & 7.47\%  & 15.76\%  & 18.22\% & 24.99\% \\    
  & 2 tok  & 15.35\% & 14.73\% & 22.16\% & 25.93\%  & 26.59\% & 24.46\% \\
  & 3 tok  & 23.18\% & 22.36\% & 26.55\% & 26.55\%  & 25.44\% & 19.82\% \\
  & 4 tok & 26.51\% & 27.94\% & 21.95\% & 17.32\%  & 15.51\% & 15.39\% \\
  & 5+ tok & 32.17\% & 31.93\% & 21.87\% & 14.44\% & 14.24\% & 15.34\% \\\hline\hline
  \multirow{3}{*}{{CANTEMIST}}
  & 1 tok  & 0.41\%  & 0.32\%  & 1.02\%  & 1.82\%  & 1.89\%  & 2.84\%  \\  
  & 2 tok  & 2.04\%  & 2.88\%  & 8.20\%  & 10.56\% & 11.88\% & 14.29\% \\
  & 3 tok  & 4.18\%  & 5.04\%  & 10.70\% & 12.79\% & 12.40\% & 11.63\% \\
  & 4 tok & 6.0\% & 7.61\% & 13.08\% & 14.31\% & 14.99\% & 15.33\% \\
  & 5+ tok & 87.37\% & 84.15\% & 67.0\% & 60.52\% & 58.84\% & 55.91\% \\\hline\hline
  \multirow{3}{*}{{ICTUSnet}}
  & 1 tok  & 1.12\% & 1.37\% & 1.70\% & 2.29\% & 3.18\% & 6.11\% \\  
  & 2 tok  & 2.37\% & 2.82\% & 4.85\% & 5.94\% & 8.12\% & 11.43\% \\
  & 3 tok & 11.36\% & 10.21\% & 15.37\% & 16.38\% & 20.90\% & 22.43\% \\
  & 4 tok & 10.04\% & 9.93\% & 13.36\% & 12.81\% & 11.08\% & 9.65\% \\
  & 5+ tok & 75.11\% & 75.67\% & 64.72\% & 62.58\% & 56.72\% & 50.4\% \\
  \bottomrule
  \end{tabular}
  \caption{Percentage of biomedical terms from the PharmaCoNER, CANTEMIST and ICTUSnet annotations that are by each model into one subword up to more than five subwords. Note that the SWM and WWM variants of the biomedical models are merged under the same names since they are equivalent for the vocabulary analysis.}
  \label{segmentations-percentages}
\end{table*}

\begin{table*}
\centering
\begin{tabular}{l|l|l|l}
\toprule
\multicolumn{1}{l|}{\textbf{Original term}} & \multicolumn{1}{l|}{\textbf{mBERT}} & \multicolumn{1}{l|}{\textbf{BETO}} & \multicolumn{1}{l}{\textbf{bio-cli-vocab-cli-52k}} \\ \hline
ADN & ADN & ADN & ADN \\
HBsAg & HB-s-A-g & H-B-s-A-g & HBsAg \\
antitransglutaminasa & anti-tra-ns-gl-uta-minas-a & anti-trans-gl-uta-minas-a & anti-trans-glutaminasa \\
clonidina & c-loni-dina & clon-idi-na & clon-idina \\
diclofenaco & di-clo-fen-aco & dic-lo-fen-aco & diclofenaco \\
glucagonoma & g-luca-gono-ma & glu-ca-gono-ma & gluca-gono-ma \\
hemangioblastomas & hem-ang-io-blast-omas & hem-ang-io-blas-tomas & hema-ngio-blas-tomas \\
hidrocortisona & hi-dro-cor-tis-ona & hidro-cor-tis-ona & hidrocortisona \\
inmunoglobulina & in-mun-og-lob-ulin-a & inmun-o-glo-bul-ina & inmun-oglobulina \\
insulina & ins-ulin-a & insulina & insulina \\
intratumorales & intrat-umo-rales & intra-tum-ora-les & intra-tum-orales  \\
leucemia & le-uce-mia & le-ucemia & leucemia \\
lidocaína & lid-oca-ína & li-doc-a-ína & lidocaína \\
linfoma & li-nfo-ma & linf-oma & linfoma \\
metastasis & meta-stas-is & metas-tasis & metastasis \\
metilfenidato & met-il-fen-idat-o & metil-fen-ida-to & metil-fenidato \\
orofaringe & oro-fari-nge & oro-far-ing-e & orofaringe \\
transaminasas & trans-ami-nas-as& trans-aminas-as & transaminasas \\
\bottomrule
\end{tabular}
\caption{Segmentation comparison between mBERT, BETO and our best model bio-cli-vocab-cli-52k. The terms are extracted randomly from the NER annotations and are presented in alphabetic order.}
\label{segmentation-examples}
\end{table*}

\section{Open Questions} \label{sec:open_questions}
In this section, we open up interesting questions motivated by the evaluation results and analysis presented in previous sections.

\paragraph{Is WWM better than SWM?}
The comparison between SWM and WWM pretraining techniques shows, in the case of the biomedical evaluation (see Table \ref{tab:biomedical_eval}), that the impact provided by the latter is affected by the vocabulary size. In particular, the WWM technique shows consistent superiority only with a vocab size of 30k. This evidence is in contrast to the finding, pointed out in \cite{pubmedbert}, that WWM is in general beneficial. We believe that further ablation studies are necessary to elucidate the interplay between the vocabulary size and the masking mechanism.

\paragraph{Is mixed-domain pretraining beneficial?}
The evaluation scores show that the bio-clinical models obtained the best performances across all tasks. Surprisingly, the bio-clinical model with clinical vocabulary obtains the highest performance on the CANTEMIST and ICTUSnet test sets. These results suggest that mixed-domain data might not always degrade performance, questioning the finding presented in \cite{pubmedbert}, where they show the negative impact of mixed-domain pretraining with biomedical and computer science text, as applied in SciBERT \cite{beltagy-etal-2019-scibert}. In our case, since we deal with two distinct but close domains,  the biomedical and the clinical ones, we somehow expect that mixed-domain pretraining could profit by adding more training data. 
On the other side, the results of the bio-cli-52k-vocab-cli model also highlight that, under the same training size conditions, the vocabulary plays an important role. In general, we believe further experiments are needed to understand the limitations of mixed-domain pretraining. However, in our evaluation scenario, we hypothesize the reason behind the encouraging performances may be related to how much overlap the specific NER data has with each model's vocabulary, as partially supported by the results obtained in Table \ref{tab:vocab_overlap_tasks}.

\section{Conclusions and Future Work}
In this work, we trained the first biomedical and clinical transformer-based pretrained language models for Spanish. We then evaluated them on a set of NER tasks, including a demanding one based on real hospital discharge reports. Our models overcome two competitive baselines, namely mBERT and BETO, representing superior solutions for biomedical NLP applications in Spanish. Finally, we analyzed the results by performing an in-depth analysis involving the model vocabularies and segmentations. Throughout the work, we outline some underexplored aspects of language model pretraining, such as the feasibility of the mixed-domain approach, the effectiveness of cross-domain transfer for clinical settings and the interplay between the vocabulary size and the token masking mechanism. Moreover, we showed the impact of different terms segmentation on the evaluation score, suggesting that over-segmentation can be detrimental for downstream tasks such as NER. Overall, we experimentally show that domain-specific pretraining has a positive impact than general-domain pretraining in a mid-resource scenario.

As future work, we suggest extending the evaluation to other tasks apart from NER, arguably the most studied task in the biomedical and clinical NLP literature but not the only relevant one. In addition, according to open questions raised in Section \ref{sec:open_questions}, we will perform more in-depth experiments to elucidate under which conditions mixed-domain pretraining is advantageous, and we will investigate the relationship between the vocabulary size and the masking mechanism.

\section{Accessibility}
We released our best models under the Apache License 2.0 to encourage the development of Spanish NLP applications in the biomedical and clinical domains. We uploaded the model in the HuggingFace models hub under the following links:
\begin{itemize}
    \item bio-52k-SWM: \url{https://huggingface.co/BSC-TeMU/roberta-base-biomedical-es}
    \item bio-cli-52k: \url{https://huggingface.co/BSC-TeMU/roberta-base-biomedical-clinical-es}
\end{itemize}
\section*{Acknowledgements}
This work was partially funded by the Spanish State Secretariat for Digitalization and Artificial Intelligence (SEDIA) within the framework of the Plan-TL and by Fundació La Marató de TV3 under the project 201712.31.
% Asier: this \clearpage command forces references to be at the end.
\clearpage

\nocite{soares-etal-2019-medical}
\bibliographystyle{plain} % We choose the &quot;plain&quot; reference style
\bibliography{paper} 
\end{document}